\documentclass{article}

\usepackage{ourArxiv}

\usepackage{listings}
\usepackage{enumitem}
\usepackage{color}
\usepackage{amsmath,amssymb,amsfonts}
\usepackage{graphicx} 
\usepackage[rightcaption]{sidecap}
\usepackage{hyperref}
\usepackage{subcaption}
\usepackage{tabularx}
\usepackage{textcomp}

\usepackage{tikz}   
\usetikzlibrary{matrix,chains,positioning,decorations.pathreplacing,arrows}
\usetikzlibrary{shapes.multipart, shapes}
\usetikzlibrary{positioning,fit,calc}
\usetikzlibrary{shadows.blur}

\title{SPFlow: An Easy and Extensible Library for Deep Probabilistic Learning using Sum-Product Networks}

\author{Alejandro Molina$^1$ \\
\texttt{molina@cs.tu-darmstadt.de} 
\And
Antonio Vergari$^2$ \\
\texttt{antonio.vergari@tuebingen.mpg.de} 
\And
Karl Stelzner$^1$ \\
\texttt{stelzner@cs.tu-darmstadt.de} 
\And
Robert Peharz$^3$ \\
\texttt{rp587@cam.ac.uk} 
\And
Pranav Subramani$^5$ \\
\texttt{p3subram@edu.uwaterloo.ca} 
\And
Nicola Di Mauro$^4$ \\
\texttt{nicola.dimauro@uniba.it} 
\And
Pascal Poupart$^5$ \\
\texttt{ppoupart@uwaterloo.ca} 
\And
Kristian Kersting$^{1,6}$ \\
\texttt{kersting@cs.tu-darmstadt.de}
\And \\
$^1$ Machine Learning Group, Computer Science Department, TU Darmstadt, Germany\\
$^2$ Probabilistic Learning Group, Empirical Inference Department, Max-Planck-Institute, Germany \\
$^3$ Machine Learning Group, Engineering Dept., University of Cambridge, UK \\
$^4$ Knowledge Acquisition \& Machine Learning Group, University of Bari ``Aldo Moro'', Italy \\
$^5$ Waterloo AI Institute, Vector Institute, University of Waterloo, CA\\
$^6$ Centre for Cognitive Science, TU Darmstadt, Germany
}

\begin{document}
\definecolor{keywords}{RGB}{255,0,90}
\definecolor{comments}{RGB}{0,0,113}
\definecolor{red}{RGB}{160,0,0}
\definecolor{green}{RGB}{0,150,0}
 
\lstset{language=Python, 
        keywordstyle=\color{keywords},
        commentstyle=\color{comments},
        stringstyle=\color{red},
        showstringspaces=false,
        identifierstyle=\color{green}}
        
\maketitle

\begin{abstract}
\noindent We introduce SPFlow, an open-source Python library providing a simple interface to inference, learning and manipulation routines for deep and tractable probabilistic models called Sum-Product Networks (SPNs). 
The library allows one to quickly create SPNs both from data and through a domain specific language (DSL). It efficiently implements several probabilistic inference routines like computing marginals, conditionals and (approximate) most probable explanations (MPEs) along with sampling as well as utilities for serializing, plotting and structure statistics on an SPN.
Moreover, many of the algorithms proposed in the literature to learn the structure and parameters of SPNs are readily available in SPFlow.
Furthermore, SPFlow is extremely extensible and customizable, allowing users to promptly distill new inference and learning routines by injecting custom code into a lightweight functional-oriented API framework.
This is achieved in SPFlow by keeping an internal Python representation of the graph structure that also enables practical compilation of an SPN into a TensorFlow graph, C, CUDA or FPGA custom code, significantly speeding-up computations.
\end{abstract}

\keywords{Sum-Product Networks \and Probabilistic Models \and Deep Learning \and Inference \and Python \and TensorFlow}

\section{Introduction}

\label{sec:introduction}

\noindent
Recent years have seen a significant interest in tractable probabilistic representations that allow exact inference on highly expressive models in polynomial time, thus overcoming the shortcomings of classical probabilistic graphical models~\cite{Koller2009}.
In particular, Sum-Product Networks (SPNs)~\cite{Poon2011}, deep probabilistic models augmenting arithmetic circuits (ACs)~\cite{Darwiche2003} with a latent variable interpretation~\cite{choiD17,Peharz2016} have been successfully employed as state-of-the-art models in many application domains such as computer vision~\cite{Gens2012,Amer2015}, speech~\cite{Zohrer2015}, natural language processing~\cite{Cheng2014,Molina2017}, and robotics \cite{Pronobis2017}.

Here, we introduce SPFlow\footnote{The latest source code and documentation are available at: \url{https://github.com/SPFlow}}, a python library intended to enable researchers on probabilistic modeling to promptly leverage efficiently implemented SPN inference routines, while at the same time allowing them to easily adapt and extend the algorithms available in the literature on SPNs.
In the following, we briefly review SPNs and some of the functions available in SPFlow. We also present a small example on how to extend the library.

\begin{SCfigure}[1.0][!t]
\begin{subfigure}{0.5\textwidth}
  \centering
  \includegraphics[height=4cm]{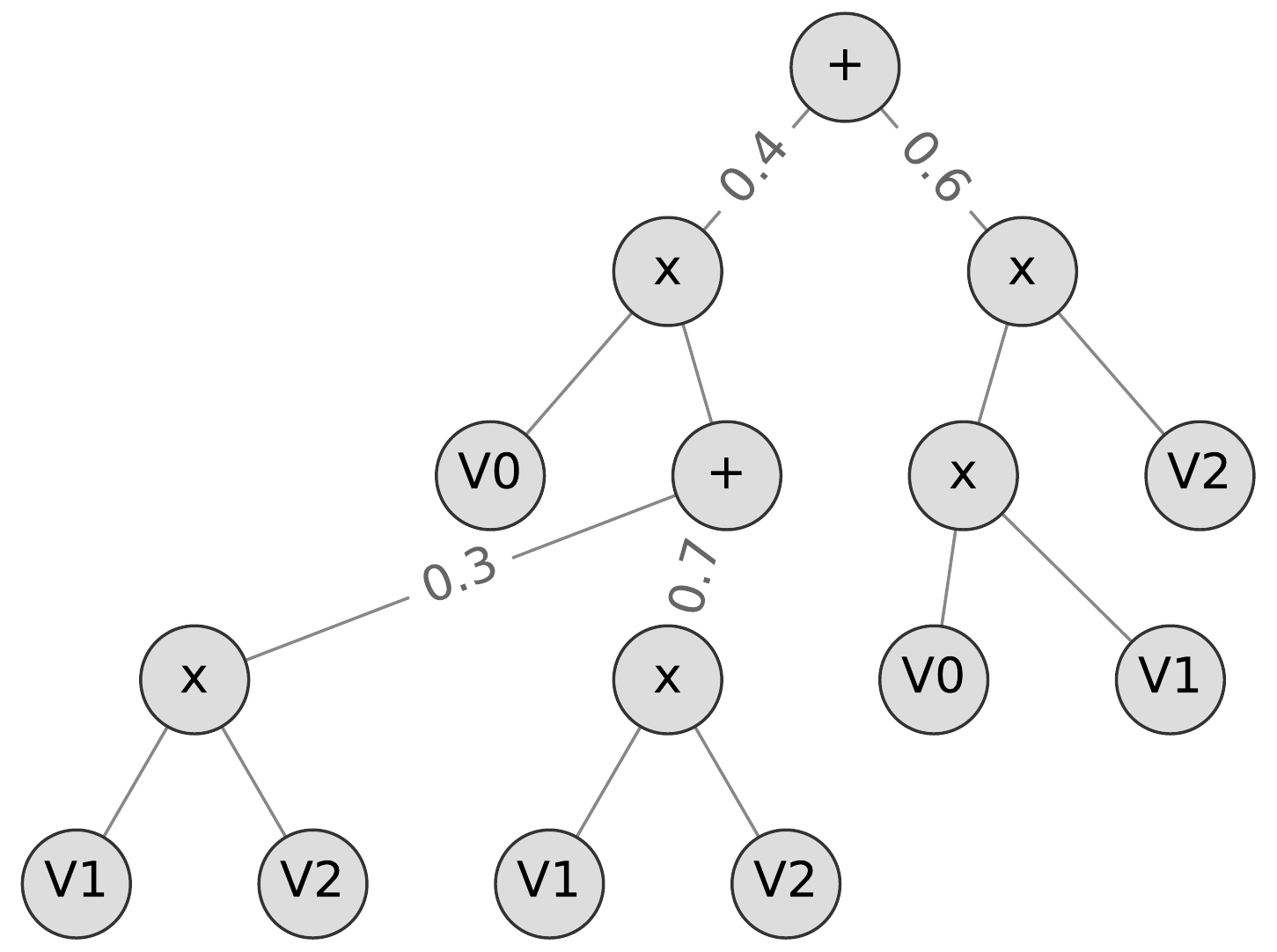}
\end{subfigure}%

\par\vspace{-10pt}
\caption{An example of a valid SPN. Here, $V_{1}$, $V_{2}$ and $V_{3}$ are random variables. The structure represents the joint distribution $P(V_{1}, V_{2},V_{3})$. }
\label{fig:spnexample}
\end{SCfigure}

\section{Sum-Product Networks} 
As illustrated in Fig.~\ref{fig:spnexample}, an SPN  is a rooted directed acyclic graph, comprising \emph{sum}, \emph{product} or \emph{leaf} nodes.
The scope of an SPN is the set of random variables appearing on the network.
An SPN can be defined recursively as follows:
(1) a tractable univariate distribution is an SPN;
(2) a product of SPNs defined over different scopes is an SPN; and
(3), a convex combination of SPNs over the same scope is an SPN. 
Thus, a product node in an SPN represents a factorization over independent distributions defined over different random variables, while a sum node stands for a mixture of distributions defined over the same variables.
From this definition, it follows that the joint distribution modeled by such an SPN is a valid probability distribution, i.e., each complete and partial evidence inference query produces a consistent probability value~\cite{Poon2011,Peharz2015a}.

To answer probabilistic queries in an SPN, we evaluate the nodes starting at the leaves. 
Given some evidence, the probability output of querying leaf distributions is propagated bottom up following the respective operations.
To compute marginals, i.e., the probability of partial configurations, we set the probability at the leaves for those variables to $1$ and then proceed as before.
To compute MPE states, we replace sum by max nodes and then evaluate the graph first with a bottom-up pass, but instead of weighted sums, we pass along the weighted maximum value.
Finally, in a top-down pass, we select the paths that lead to the maximum value, finding approximate MPE states \cite{Poon2011}.
All these operations traverse the tree at most twice and therefore can be achieved in linear time w.r.t. the size of the SPN.

\section{An Overview of the SPFlow Library}
As most operations on SPNs are based on traversing the graph in a bottom-up or top-down fashion, we model the library as basic node structures and generic traversal operations on them. The rest of the SPN operations are then implemented as lambda functions that rely on the generic operations. 

Therefore, the SPFlow library puts the graph structure at the center. All other operations receive or produce a graph that can be then used by the other operations. This increases the flexibility and potential uses. As an example, one can create a structure using different algorithms and then save it to disk. Later on, one can load it again  and do parameter optimization using, e.g.,  TensorFlow~\cite{tensorflow2015-whitepaper}, and then do inference to answer probabilistic queries. All those operations can be composed as they rely only on the given structure. More specifically, the functionality of SPFlow covers:

\begin{center}
\begin{tabularx}{1\linewidth}{  X  X  X  X } \hline 
\bf Modeling  & \bf Evaluation & \bf Sampling & \bf Other \\ 
\small
\begin{itemize}[noitemsep,topsep=0pt, leftmargin=*]
\item Domain specific\newline language (DSL)
\item Structure \mbox{Learning}
\item Random \mbox{Structures}
\item Checks for\newline \mbox{consistency} and\newline \mbox{completeness}
\end{itemize} & 
\small
\begin{itemize}[noitemsep,topsep=0pt, leftmargin=*]
\item Joint queries
\item Marginal queries
\item Approximate most probable\newline explanation (MPE)
\item Parameter\newline Optimization
\end{itemize} & 
\small
\begin{itemize}[noitemsep,topsep=0pt, leftmargin=*]
\item Ancestral \mbox{Sampling}
\item Conditional \mbox{Sampling}
\end{itemize} &
\small
\begin{itemize}[noitemsep,topsep=0pt, leftmargin=*]
\item Plotting
\item JSON
\item Text Formats
\item Standard Pickling
\item \mbox{Tensorflow} Graphs
\item Convert to C code
\end{itemize}
\\
\hline 
\end{tabularx}
\end{center}

\newpage

\section{SPFlow Programming Examples}
To create the SPN shown already in Fig.~\ref{fig:spnexample}, one simply
writes the follow code after loading the library:

{\scriptsize \begin{lstlisting}[language=Python]
spn = 0.4 * (Categorical(p=[0.2, 0.8], scope=0) * 
             (0.3 * (Categorical(p=[0.3, 0.7], scope=1) * 
                     Categorical(p=[0.4, 0.6], scope=2)) 
            + 0.7 * (Categorical(p=[0.5, 0.5], scope=1) * 
                     Categorical(p=[0.6, 0.4], scope=2)))) 
    + 0.6 * (Categorical(p=[0.2, 0.8], scope=0) * 
             Categorical(p=[0.3, 0.7], scope=1) * 
             Categorical(p=[0.4, 0.6], scope=2))
\end{lstlisting}}

Alternatively, we can create the same spn using the following code:
{\scriptsize \begin{lstlisting}[language=Python]
p0 = Product(children=[Categorical(p=[0.3, 0.7], scope=1), Categorical(p=[0.4, 0.6], scope=2)])
p1 = Product(children=[Categorical(p=[0.5, 0.5], scope=1), Categorical(p=[0.6, 0.4], scope=2)])
s1 = Sum(weights=[0.3, 0.7], children=[p0, p1])
p2 = Product(children=[Categorical(p=[0.2, 0.8], scope=0), s1])
p3 = Product(children=[Categorical(p=[0.2, 0.8], scope=0), Categorical(p=[0.3, 0.7], scope=1)])
p4 = Product(children=[p3, Categorical(p=[0.4, 0.6], scope=2)])
spn = Sum(weights=[0.4, 0.6], children=[p2, p4])

assign_ids(spn)
rebuild_scopes_bottom_up(spn)
\end{lstlisting}}

\noindent Actually, Fig.~\ref{fig:spnexample} itself was plotted by calling
\lstinline[language=Python]{plot_spn(spn, 'basicspn.pdf')}. To evaluate the likelihood of the SPN on some data, one can use Python or TensorFlow:

{\scriptsize \begin{lstlisting}[language=Python]
test_data = np.array([1.0, 0.0, 1.0]).reshape(-1, 3)
log_likelihood(spn, test_data) % [[-1.90730501]]
eval_tf(spn, test_data) % [[-1.90730501]]
\end{lstlisting}}

\noindent To learn the parameters of the SPN using TensorFlow, one calls
{\scriptsize \begin{lstlisting}[language=Python]
optimized_spn = optimize_tf(spn, test_data)
log_likelihood(optimized_spn, test_data) % [[-1.38152628]]
\end{lstlisting}}

\noindent Marginal likelihoods just require setting "nan" on the features to be marginalized:
{\scriptsize \begin{lstlisting}[language=Python]
log_likelihood(spn, np.array([1, 0, np.nan]).reshape(-1, 3) % [[-1.2559681]]
\end{lstlisting}}

\noindent Sampling creates instances where samples are obtained for the cells that contain "nan": 
{\scriptsize \begin{lstlisting}[language=Python]
sample_instances(spn, np.array([np.nan, 0, 0]).reshape(-1, 3), RandomState(123))
%[[1. 0. 0.]]
\end{lstlisting}}

\noindent To learn the structure of an SPN, say for binary classification, let us first create a 2D dataset with a binary label. An instance has label 0, when the features are close to the generating Gaussian with mean 5. It has label 1, when the features are closer to the generating Gaussian with a mean of 15:

{\scriptsize \begin{lstlisting}[language=Python]
train_data = np.c_[np.r_[np.random.normal(5, 1, (500, 2)), 
                        np.random.normal(15, 1, (500, 2))],
                   np.r_[np.zeros((500, 1)), np.ones((500, 1))]]
\end{lstlisting}}

\noindent Now we specify the statistical types of the random variables and learn a SPN classifier:

{\scriptsize \begin{lstlisting}[language=Python]
ds_context = Context(parametric_type=[Gaussian, Gaussian, Categorical])
ds_context.add_domains(train_data)
spn = learn_classifier(train_data,ds_context,learn_parametric, 2)
\end{lstlisting}}

\noindent Doing MPE on the classification SPN gives us the classifications.

{\scriptsize \begin{lstlisting}[language=Python]
mpe(spn, np.array([3.0, 4.0, np.nan, 12.0, 18.0, np.nan]).reshape(-1, 3))
% [[ 3.  4.  0.]
%  [12. 18.  1.]]
\end{lstlisting}}

\noindent The third column is the label and we can see that it behaves as expected in this synthetic example.

\newpage

\section{Extending the SPFlow library}
To illustrate the flexibility of SPflow, we show how to extend inference to other leave types. Here we implement the Pareto leaf distribution. It relies on the infrastructure already present. 
{\scriptsize \begin{lstlisting}[language=Python]
class Pareto(Leaf):
    def __init__(self, a, scope=None):
        Leaf.__init__(self, scope=scope)
        self.a = a

def pareto_likelihood(node, data, dtype=np.float64):
    probs = np.ones((data.shape[0], 1), dtype=dtype)
    from scipy.stats import pareto
    probs[:] = pareto.pdf(data[:, node.scope], node.a)
    return probs

add_node_likelihood(Pareto, pareto_likelihood)

spn =  0.3 * Pareto(2.0, scope=0) + 0.7 * Pareto(3.0, scope=0)

log_likelihood(spn, np.array([1.5]).reshape(-1, 1))
%[[-0.52324814]]
\end{lstlisting}}

\noindent The same kind of extensions are possible for all other operations. This way it is easy to extend the library, by adding new nodes or even new operations. For instance, one could easily interface probabilistic programming languages and tools such as PyMC or Pyro.

{\bf Acknowledgements.} {RP acknowledges support from the European Union's Horizon 2020 research and innovation programme under the Marie Sk\l{}odowska-Curie Grant Agreement No.~797223 --- HYBSPN. This work has benefited from the DFG project
CAML (KE 1686/3-1), as part of the SPP 1999, and from
the BMBF project MADESI (01IS18043B).}

\vskip 0.2in
\bibliographystyle{unsrt}
\bibliography{references}

\end{document}